\documentclass{article}
\usepackage{spconf,amsmath,graphicx}

\usepackage{enumitem}
\setlist{nosep, leftmargin=14pt}

\usepackage{etoolbox}

\usepackage{caption}
\makeatletter
\let\oldthebibliography\thebibliography
\renewcommand{\thebibliography}[1]{%
  \oldthebibliography{#1}%
  \setlength{\itemsep}{9pt}%
  \setlength{\parskip}{0pt}%
  \setlength{\parsep}{0pt}%
}
\makeatother

\usepackage{mwe} 
\usepackage{setspace}

\title{BrainSegNet: A Novel Framework for Whole-Brain MRI Parcellation Enhanced by Large Models}
\name{Yucheng Li, Xiaofan Wang, Junyi Wang ,Yijie Li, Xi Zhu, Mubai Du, Dian Sheng, Wei Zhang, Fan Zhang$^{\star}$\thanks{$^{\star}$Corresponding author: Fan Zhang, fan.zhang@uestc.edu.cn}}
\address{University of Electronic Science and Technology of China, Chengdu, China}
\begin{document}
%
\maketitle
\begin{abstract}
Whole-brain parcellation from MRI is a critical yet challenging task due to the complexity of subdividing the brain into numerous small, irregular shaped regions. Traditionally, template-registration methods were used, but recent advances have shifted to deep learning for faster workflows. While large models like the Segment Anything Model (SAM) offer transferable feature representations, they are not tailored for the high precision required in brain parcellation. To address this, we propose BrainSegNet, a novel framework that adapts SAM for accurate whole-brain parcellation into 95 regions. We enhance SAM by integrating U-Net skip connections and specialized modules into its encoder and decoder, enabling fine-grained anatomical precision. Key components include a hybrid encoder combining U-Net skip connections with SAM’s transformer blocks, a multi-scale attention decoder with pyramid pooling for varying-sized structures, and a boundary refinement module to sharpen edges. Experimental results on the Human Connectome Project (HCP) dataset demonstrate that BrainSegNet outperforms several state-of-the-art methods, achieving higher accuracy and robustness in complex, multi-label parcellation.
\end{abstract}
\begin{keywords}
Magnetic Resonance Imaging, Brain parcellation, Deep Learning
\end{keywords}
\section{Introduction}
\label{sec:intro}

Whole‑brain parcellation from MRI refers to the process of partitioning a complete brain image volume into a comprehensive set of anatomically meaningful regions of interest (ROIs), including cortical and sub‑cortical structures, white matter, cerebrospinal fluid, and other tissue types — effectively assigning each voxel in the brain scan to one of many predefined labels \cite{1,2}. In clinical workflows and large‑scale neuroimaging studies, the ability to generate accurate, reproducible whole‑brain parcellations rapidly enables quantification of brain morphology, assessment of disease‑related structural change, surgical planning, and longitudinal monitoring \cite{4,5}. Currently, the mainstream tool for MR parcellation is template-based approaches such as FreeSurfer \cite{26}. However, a major limitation of such approaches is its computational inefficiency which can take hours to complete.

Recent advances in deep learning largely improved brain parcellation tasks. Traditional convolutional encoder–decoder models such as U‑Net have long served as foundational tools in medical imaging parcellation \cite{6,7,8}, including brain MRI tasks \cite{9}, by virtue of their symmetric contraction‑expansion pathways and skip‑connections which enable fine localization. Indeed, U‑Net variants and their 3D extensions have shown strong accuracy in organ‑specific and lesion‑specific parcellation tasks \cite{27,28}. However, when applied to the full brain parcellation case—with dozens of cortical and sub‑cortical labels, varying structure sizes—these models exhibit reduced accuracy. For instance, representative learning-based MRI parcellation pipelines like FastSurfer \cite{9} using a U-Net have demonstrated high speed and accuracy in whole-brain parcellation, while their performance on small brain regions are still limited. 
	
Recently, large models such as the Segment Anything Model (SAM) \cite{13} and its variants (e.g., SAM2) \cite{14} have emerged as powerful tools for general‑purpose image parcellation. These models are distinguished by their training on massive natural‑image datasets, which endows them with highly transferable feature extraction capabilities and the flexibility to be prompt‑driven (e.g., via points or bounding boxes) for downstream parcellation tasks. These characteristics make them promising candidates for medical image computing, a field often challenged by scarce annotated data, diverse imaging modalities, and the impracticality of re-training models for every specific clinical task. However, applying these models to the task of multi-label brain MRI parcellation—which involves segmenting small cortical and sub-cortical structures, thin ribbons, and subtle intensity differences between tissue boundaries—reveals several key limitations \cite{15,16,17}. First, since SAM training primarily uses 2D natural images, the difference between 2D natural image data and 3D low-contrast medical scan data weakens the original SAM model's parcellation ability in medicine. For example, SAM performance varies significantly depending on the target anatomical structure, performing worse on tasks involving weak or blurred boundaries \cite{18}. Second, the human brain is a complex structure, requiring models to possess volumetric contextual information and domain-specific adaptive capabilities. However, the original SAM model lacks these inherent mechanisms and is therefore unsuitable for such sophisticated three-dimensional MRI parcellation tasks. Third, while medical variants of SAM have shown breakthroughs in tasks such as tumor and liver parcellation \cite{19,21}, there is still no large‑scale model dedicated to parcellating the entire brain. Therefore, we propose to leverage the SAM model for the task of high-fidelity, multi-label whole-brain MRI parcellation.

In this work, we propose BrainSegNet, a novel deep learning framework that adapts and enhances SAM for accurate whole-brain MRI parcellation into 95 regions. The network integrates the non-cue-driven capabilities of SAM with the hierarchical, multi-scale decoder with multiple detail-oriented modules. We design a SAM-Based U-Net encoder that combines the long skip connection idea of  U-Net with the transformer block of SAM to preserve fine-grained spatial details. We also design a novel multi‑scale attention decoder, that incorporates three novel design elements: 

(1) An Atrous Spatial Pyramid Pooling (ASPP) module to capture contextual information at multiple dilation rates and global pooling for improved parcellation of varied‑size structures; 

(2) A Channel and Spatial Attention (CSA) to adaptively re‑weight channel and spatial features, boosting sensitivity to small, irregular shaped anatomical regions; 

(3) A Boundary Refinement (BF) module that injects explicit edge cues to sharpen interfaces and mitigate boundary leakage. 

Overall, we provide a fully integrated model‑level solution that delivers robust performance—demonstrating consistent parcellation accuracy of 95 brain regions without the need for region‑specific tuning or post‑hoc corrections. Our experimental evaluation demonstrates that our framework significantly outperforms both traditional U‑Net variants and SAM‑only pipelines, offering a robust solution for automatic whole‑brain parcellation.

\section{Methodology}
\label{sec:method}
\subsection{Method Overview}

\begin{figure}[htb]
    \centering
    \includegraphics[width=0.48\textwidth]{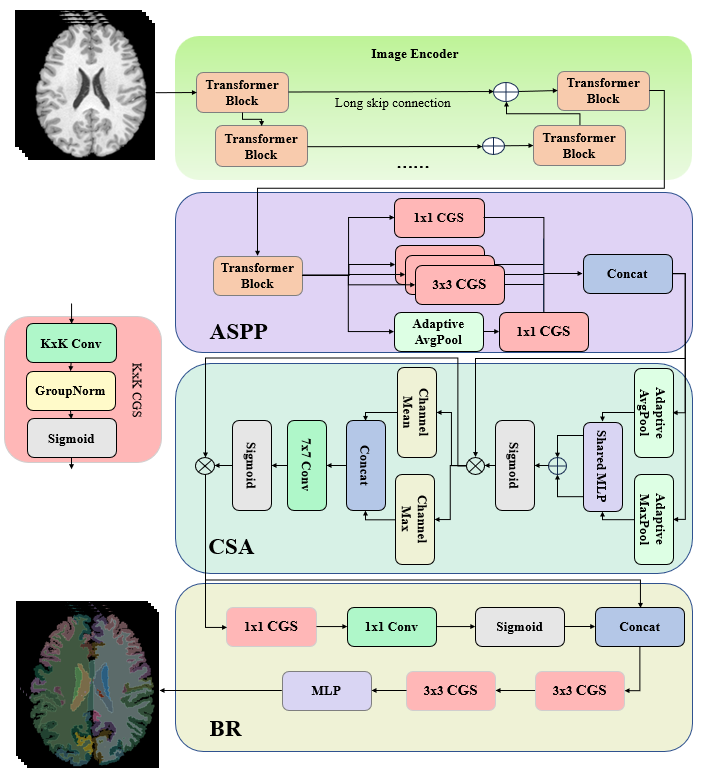}  
  \caption{Overview of the proposed BrainSegNet framework for whole-brain multi-label parcellation.}  
  \label{fig:arch} 
\end{figure}

\textbf{Figure.}\ref{fig:arch} gives a method overview, with three major sequential stages. First, the input MRI images are passed into the symmetric skip‑connected U‑Net encoder, where the ASPP module extracts multi‑scale contextual feature maps. Next, these features are refined by the CSA attention module, which re‑weights channels and spatial regions to highlight subtle anatomical structures and suppress irrelevant responses. Finally, the boundary‑refinement head takes the attention‑enhanced features and predicts an auxiliary edge map, which is used to sharpen boundaries and reduce label leakage between adjacent regions. Overall, this pipeline integrates feature extraction, focused attention, and precise boundary correction to achieve robust whole‑brain multiregion parcellation.

\subsection{SAM-Based U-Net Encoder}

The SAM encoders consist of 12 pre-trained transformer blocks \cite{20}. While they perform well in parcellating natural images, they are not trained specifically for medical images, and their excessive depth can potentially cause the loss of details in medical images. On the other hand, U-Net has advantages in medical parcellation tasks, with skip connections to effectively capture image details. Therefore, to maximize the preservation of brain image information, we symmetrically connect the 12 transformer blocks of SAM using skip connections. To adapt the encoder to medical images, we use the 3D adapter and low-rank factor in MASAM to fine-tune the ViT blocks in the encoder \cite{21}. All transformer blocks were fine-tuned.

\subsection{Multi‑Scale Attention Decoder}

We integrate the ASPP, CAS and BF modules into a unified pipeline as the decoder for parcellation. The following is an introduction to each module of decoder.

The ASPP module is incorporated to capture rich multi‑scale context from decoder feature maps. Inspired by work in \cite{22}, the module applies multiple parallel dilated (atrous) convolutions at varying dilation rates—e.g., rates of 1, 6, 12, 18—so that each branch probes a different effective receptive field without down‑sampling the spatial resolution.Unlike traditional ASPP, we design a dual ASPP that incorporates global pooling before the bottom convolution. A global‑pooling branch aggregates full‑feature‑map context which is then up‑sampled and concatenated with the dilated‑conv outputs. These parallel streams are concatenated and projected via a 1$\times$1 convolution to fuse the multi‑scale cues into a unified feature tensor. This design helps the model recognize anatomical structures at very different scales—from large lobes to tiny nuclei—while maintaining high spatial resolution. 

The CSA module is applied to refine and re‑weight feature responses so that informative channels (feature maps) and spatial locations (within each map) become more prominent. Drawing on the overall framework ideas in \cite{23}, we design a CSA module that utilizes global and channel pooling concatenation.Concretely, the channel‑attention sub‑block begins by computing both global average‑pooling and global max‑pooling per channel (thus squeezing spatial dimensions), feeds the resulting vectors into a shared multi‑layer perceptron (MLP) and applies a sigmoid gate. The resulting channel‑attention weights are used to scale the original feature maps. Next, the spatial‑attention sub‑block takes the channel‑weighted feature maps, applies pooling across the channel dimension, concatenates them, passes through a convolution and sigmoid activation to generate a spatial‑attention map. This map highlights “where” in the spatial plane the model should focus. Together, CSA ensures that fine structures (e.g., small sub‑cortical nuclei) are boosted in representation, while noise or background clutter is suppressed, thereby improving sensitivity to small, irregular shaped anatomical regions. 

The BF module  is stacked on top of the decoder output to explicitly refine interfaces between adjacent anatomical regions—where parcellation errors often occur due to blurred boundaries or partial‑volume effects. Influenced by the work of \cite{24}, we design a similar offset branch. The head takes the decoder’s up‑sampled feature tensor, predicts an auxiliary edge map (usually via shallow convolution layers followed by a sigmoid). The edge map is concatenated back with the decoder features (or added) and processed with a few shallow convolutions (e.g., 3×3, GroupNorm, ReLU) to produce a final refined parcellation output. This mechanism sharpens boundaries and limits “bleeding” of labels across adjacent structures. Using edge guidance in parcellation has been shown to improve contour accuracy and reduce boundary error in medical volumes.

\subsection{Model Training and Testing}
The fine‑tuning process of BrainSegNet is guided by a hybrid parcellation loss, defined as $L_{seg}$ = $\alpha$ $L_{ce}$ + $\beta$ $L_{Dice}$. Weighting factors $\alpha$ and $\beta$ are set to 0.2 and 0.8, respectively. The network receives five consecutive slices as input to capture local volumetric context. To enhance the generalization ability of the model during fine‑tuning, we apply flipping and random Gaussian noise. Training uses the Adam optimizer with a batch size of 8. A warm‑up strategy is adopted: the learning rate is linearly increased to the target value and then decayed exponentially to stabilize convergence. We use a ViT‑B backbone for the image encoder and train over 70 epochs. Implementation is in PyTorch 2.0 on a single NVIDIA A100 GPU.

\begin{figure*}[th]
  \centering
  \includegraphics[width=\textwidth]{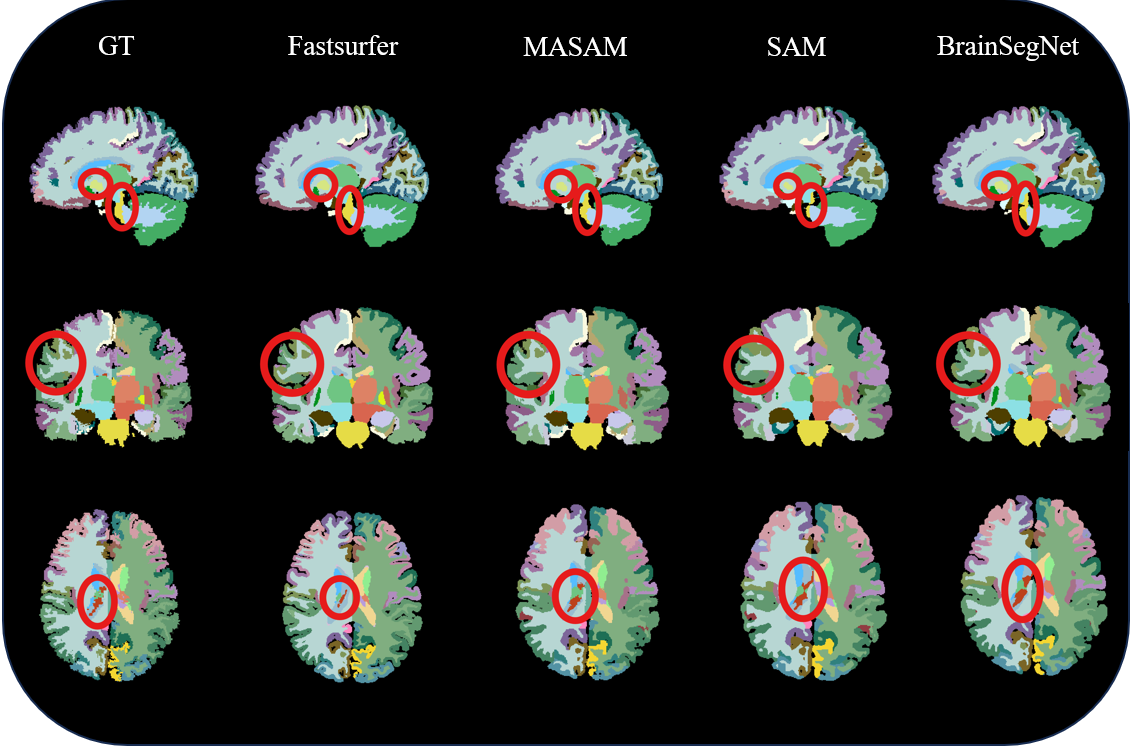} 
  \caption{Visualization of a representative sagittal slice: GT (ground truth), FastSurfer, MASAM, SAM, and BrainSegNet. The regions circled in red indicate statistically significant differences.}
  \label{fig:brain}
\end{figure*}

\subsection{Datasets}
We use the Human Connectome Project (HCP) \cite{25} MRI dataset acquired from healthy adults. There are 100 T1 subjects, including 90 subjects for training and 10 for testing.The voxel size is 0.7 mm, and the resolution is 260$\times$311$\times$260. To adapt to VIT, we crop all data to 256$\times$256$\times$260. A total of 95 labeled regions from the entire brain were used,  derived from FreeSurfer \cite{26} software.

\begin{table}[htb]
  \centering
  \caption{Quantitative results on the HCP testing subjects across 95 brain regions. Top: comparison with state-of-the-art methods (FastSurfer, MASAM, SAM). Bottom: ablation of model components (U-Net, ASPP, CSA, and BR).}  
  \label{fig:table}  
  \includegraphics[width=0.48\textwidth]{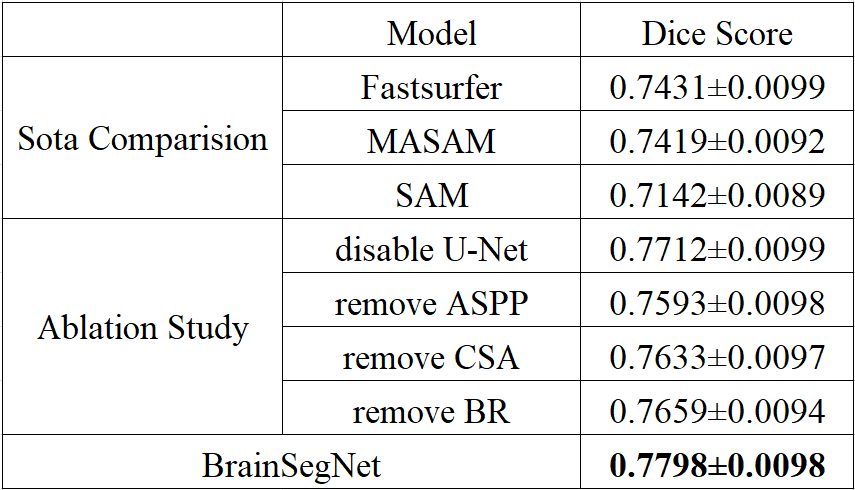}  
\end{table}

\section{Result}
We compare BrainSegNet with several state-of-the-art (SOTA) methods including FastSurfer \cite{9}, MASAM \cite{21}, and SAM \cite{13}. For Fastsurfer, it is designed to perform whole-brain parcellation, so we directly use its GitHub version for inference. For MASAM, we remove the complex  enhancement processes from its preprocessing, and replace them with our random flipping and random Gaussian noise. For SAM, we use the original SAM model and fine-tuned it using the MASAM fine-tuning method.

We perform ablation studies by removing the U-Net skip connection in the encoder, and CSA, ASPP, or BR from the decoder, while keeping others unchanged. That is, in the remove-CSA model, the ASPP output is the input of the BR module, , bypassing CSA. Similarly, in the remove-ASPP model, the CSA input is directly connected to the image encoder.For the remove-BR model, the output of the CSA module is directly connected to a MLP. For the disabled-U-Net model, we remove the long skip connection from the image encoder. Other training and testing parameter settings are consistent with the full BrainSegNet in the SOTA comparison. All evaluations are performed on the HCP testing data, and the mean Dice (averaged over classes and subjects) are used as the evaluation metric.

\textbf{Table} \ref{fig:table} presents a comparison of Dice scores across several parcellation models, including SOTA models and an ablation study of the proposed BrainSegNet framework. In the SOTA comparison section, FastSurfer achieves the highest Dice score of 0.743, followed closely by MASAM at 0.741, and SAM at 0.714. These results highlight the strong performance of traditional parcellation models, but also reveal their limitations when compared to more specialized architectures. In the ablation study, the performance of BrainSegNet shows that each individual component contributes positively to the overall parcellation accuracy. The full BrainSegNet model achieves the highest Dice score of 0.779, demonstrating that combining ASPP, CSA, and boundary refinement leads to significant improvements in parcellation accuracy, especially in complex and small anatomical regions.

\textbf{Figure} \ref{fig:brain} presents a visual comparison of whole-brain parcellation results from multiple models, including the ground truth (GT), FastSurfer, MASAM, SAM, and BrainSegNet. The parcellation results are shown in both sagittal and coronal views of the brain, with various brain regions represented by distinct colors. The circled regions highlight areas where parcellation discrepancies occur, particularly in smaller or irregular structures. As observed, while BrainSegNet achieves more accurate delineation in these challenging areas, models such as FastSurfer and SAM show noticeable misclassifications. These results underscore the advantage of the proposed framework in providing more precise parcellation, particularly for small and anatomically complex regions, as compared to other baseline models.

\section{Conclusion}
In this study, we propose BrainSegNet, a novel deep learning framework for whole-brain parcellation, with successful application on dataset. BrainSegNet preserves significant spatial features, and enables discriminative representations of brain regions. Compared to the SOTA methods, BrainSegNet is the top performer on a ground truth labeled dataset acquired testing datasets. BrainSegNet enables effective whole-brain parcellation.

\section{Compliance With Ethical Standards}
This study was conducted retrospectively using public HCP imaging data. No ethical approval was required.
\section{Acknowledgements}
This work is in part supported by the National Key R\&D Program of China (No. 2023YFE0118600) and the National Natural Science Foundation of China (No. 62371107).

\begingroup
\setstretch{0.75}   
\bibliographystyle{IEEEbib}
\bibliography{ref}
\endgroup

\end{document}